# EFFICIENT DEEP LEARNING METHODS FOR IDENTIFICATION OF DEFECTIVE CASTING PRODUCTS

Bharath Kumar Bolla[1], [*] Mohan Kingam[2,4], Sabeesh Ethiraj[3]

[1]Salesforce, Hyderabad, India
[2]bolla111@gmail.com

[2]Upgrad Education Pvt. Ltd., Mumbai, India
[1]mohan.kingam@gmail.com

[3]Command Hospital, Bangalore, India
[3]sabeesh90@yahoo.co.uk

[4]Hexagon Capability Center India Pvt. Ltd., Hyderabad, India
[4]mohan.kingam@hexagon.com

**Abstract.** Quality inspection has become crucial in any large-scale manufacturing industry recently. In order to reduce human error, it has become imperative to use efficient and low computational AI algorithms to identify such defective products. In this paper, we have compared and contrasted various pre-trained and custom-built architectures using model size, performance and CPU latency in the detection of defective casting products. Our results show that custom architectures are efficient than pre-trained mobile architectures. Moreover, custom models perform 6 to 9 times faster than lightweight models such as MobileNetV2 and NasNet. The number of training parameters and the model size of the custom architectures is significantly lower (~386 times & ~119 times respectively) than the best performing models such as MobileNetV2 and NasNet. Augmentation experimentations have also been carried out on the custom architectures to make the models more robust and generalizable. Our work sheds light on the efficiency of these custom-built architectures for deployment on Edge and IoT devices and that transfer learning models may not always be ideal. Instead, they should be specific to the kind of dataset and the classification problem at hand.

# 1    Introduction

Quality control is vital in many industries, especially those that use casting or welding. Product quality affects customer satisfaction and loyalty. Inspection and testing are vital parts of the manufacturing process because they help control quality, reduce costs, prevent the loss, and locate defects. While most of these flaws are detectable with the naked eye, human inspection is time consuming, error prone, costly, and unreliable. Automated visual inspection solutions are helping companies overcome these obstacles. Most of the manufacturing sectors have relied on various non-destructive methodologies such as ultrasonic testing, magnetic particulate control and real time. X-ray image analysis started to gain popularity before deep learning methods like Convolution Neural Networks. Technological advancements in high-resolution X-rays have increased the detection capacity through 3D-characterization [1]. Gabor filters are another popular defect detection method. The image can be decomposed into distinct components based on scale and orientation. They are widely used in defect detection because they provide the most precise spatial localization [2].

Despite their efficiency and robustness, deep learning-based models are difficult to deploy on devices with limited memory, such as smartphones, tablets, and IoT devices. In the industrial setting, hosting deep learning models in the cloud is impractical due to latency and maintenance costs. The paper aims to evaluate both custom models and transfer learning architectures to identify the best performing model in evaluation metrics such as Accuracy, Recall, F1 scores, and model size. Inference time on different sizes of datasets will also be calculated to identify faster performing models suitable for deployment on IoT and Edge devices. Further, augmentation techniques will be evaluated on the hypothesized best performing custom model to establish their robustness and generalizability on augmented datasets.

# 2    Literature Review

Computer vision is used to check for defects in different manufacturing products made from steel, aluminuim, glass, fabric and polycrystalline materials [3]. Vision-based defect detection detects internal flaws in aluminum alloy castings in addition to external flaws. The defects can be seen in X-ray images of the affected components, such as brake drums, gears, and the engine body. Combination of Deep learning and X-ray images can be used to detect internal flaws in aluminum casting parts [4].

With deep learning-based image tasks outperforming the average human inspection, automated vision inspection systems for inspecting surface defects in casting products [5] are becoming more common. AlexNet to MobileNet, Deep Neural Networks have improved accuracy, decreased model parameters, the total number of operations (flops), memory footprint, and computation time over the years significantly. Accuracy as a function of parameter count, also known as information density, is a performance metric that emphasizes a particular architecture's breadth to maximize its parametric space utilization. This accuracy function revealed that basic models like VGG and AlexNet are larger because they have not fully exploited their learning capability. In contrast, more efficient models like ResNet, GoogLeNet, and ENet have higher accuracy per parameter by training all neurons on the given task [6].

Convolution Neural Networks are known for their ability to extract features. The image's representation is learned by the convolutional layers, which can then be used for classification, object detection, and recognition. Due to the difficulty of obtaining a large enough dataset to make the model robust enough to be reused in any type of image classification problem, we also train an entire CNN from scratch on a very rare occasion [7]. The concept of transfer learning entails the use of weighted pre-trained networks. Deep neural networks tend to overfit the training dataset because they are complex networks with a large number of parameters.

To make a more compact representation, CNN uses the pooling layer. Pooling reduces feature map height and width and reduces the parameter count [8]. The most common pooling methods are max, average, and global average. The Max Pooling layer reduces the output from the previous layer by selecting the maximum value in each feature map. Thus, trying to extract the image's dominant feature. The pooling layer considers a feature detected if any of the patches strongly believe it exists [9]. The Average pooling layer, on the other hand, takes the mean of all the weighted values extracted, to determine the most prominent feature. Max pooling is more popular than average pooling because it performs better as it ignores minor changes by taking away the location flaws in the features [10]. Global average pooling is an alternative to fully connected layers for pooling. Global average pooling can be applied to feature maps to avoid overfitting and to make the model generic. It allows the output layer to get the average vector from each feature map in the final convolution layer, making the process more network-centric and aligned with the output classification categories [11].

Data augmentation can be used to strengthen the model and compensate for class imbalances. Techniques for enhancing data include flipping, rotating, and zooming. Affine transformation shears the image while keeping the other vector constant. This creates synthetic data and improves model robustness during training [12].

Most deep convolutional models have large parameters and are designed to improve accuracy. However, these aren't readily suitable for edge or mobile devices. A new class of efficient models have evolved for mobile and embedded vision applications which are known as MobileNets [13]. These networks are mobile based models which focuses on reducing the number of operations and the latency of the model. MobileNets use depthwise separable convolutions to build light weight deep neural networks. Another compressed network called NASNet is a mobile network built with depth-dependent and grouped convolutions. Grouped convolution uses parallel processing by splitting the filters into two groups, one for each input depth [14]. ResNet50 is another network which has proved that the complexity of the network can be decreased even when more layers are added to it by training the model on residuals. Other than mobile networks, ResNet50 is also a popular network that is widely used producing the best compact model with high accuracy [15]. Straightforward transfer learning in Deep Learning should not be implemented and it requires layer fine tuning [16].

not be implemented Research Methodology

## 2.1 Dataset description

The dataset consists of images depicting the front view of an impeller casting from a castings manufacturing company. These images are RGB images consisting of three channels and are divided into two folders consisting of train and test images of two classes (Normal and Defective). The number of train and test images are 6633 and 715 respectively of size 300x300x3 pixels. Usig Image generators, all the images are split into train, validation, and test containing 5307, 1326, and 715 images respectively.

## 2.2 Data Preprocessing - Reducing training parameters

The image data consists of three channels. The number of training parameters is reduced by converting the image from RGB to grayscale (3 to 1 channel). This conversion is done only with custom architectures as transfer learning architectures such as Resnet, MobileNetV2, and NasNet require the input image to contain three channels. The images are scaled using appropriate scaling techniques to ensure no over representation of a particular set of pixels during model training.

## 2.3    Creation of train and customized test generators

The train generator and test generators are created using Tensorflow's ImageDataGenerator. Two different sets of train and test generators are created, one each for with and without augmentation. Five additional test generators are created with augmented and non-augmented datasets, each with different batch sizes (1,10,50,100,715). This is done to calculate the inference time of the model in predicting the different number of images.

## 2.4    Data Augmentation

Data augmentation techniques such as ZCA Whitening, Flipping (Horizontal and Vertical), Rotating, and Zooming have been used to enhance the model's robustness. Both standard and augmented test datasets were used to evaluate the effects of these techniques on the model's overall performance.

## 2.5    Calculation of Inference Timings

Inference time for a specified number of images is calculated using the customized test dataset with different batch sizes mentioned in Equation-1.

$$Inference\ time\ = \frac{Inference\ time\ of\ the\ Total\ test\ dataset}{Number\ of\ batches\ in\ the\ test\ dataset}$$

*Equation 1. Inference Time Calculation*

## 2.6    Model Size Reduction: Channel Pruning, GAP, Parameter Tuning

Three transfer learning models as MobileNetV2, NasNet, and Resnet50 and custom model using augmentation and without augmentations, have been built to evaluate the effect of these models on the inference time.

**Paramater Tuning.** The custom model was built to reduce the number of training parameters. The input image is converted to a monochrome grayscale image, thereby resulting in the reduction of training parameters in the first convalution layer of the deep network.

**Channel Pruning**. The concept of channel pruning was used to achieve model compression by sequentially reducing the number of output channels over successive convolution layers without compromising on the model performance. Further, the output of the last convolutional layer is a single neuron sigmoid function layer, as opposed to a two neuron output in other models.

**Global Average Pooling(GAP).** GAP is added before the soft max layer to reduce the number of neurons. Fig. 1 depicts the above mentioned methodologies.

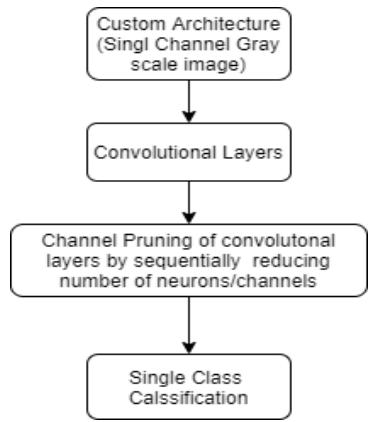

*Figure 1. Custom model flow diagram*

### 2.7     Pre-Trained Architectures - ResNet50, MobileNetV2 & NasNet

The transfer learning architectures used here are trained using their original weights used in the ImageNet Classification. The network's top layer is replaced with a ***two-neuron custom softmax layer*** with ***binary Cross Entropy*** as the loss function. The input image to the architecture is an ***RGB*** image with a ***three-channel*** dimension, unlike the custom architecture, as these networks are pre-trained only on three-channel images. The architectures are shown in Fig 2, Fig3, and Fig 4.

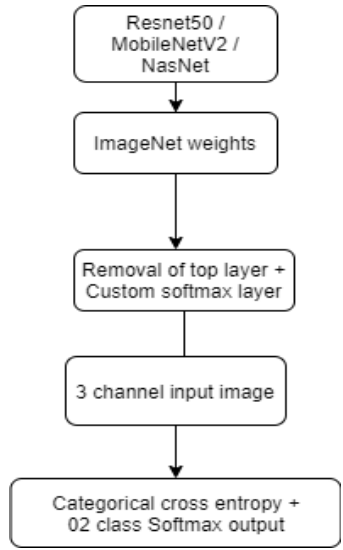

*Figure 2. Transfer Learning Flow*

## 2.8    Loss Function

The loss function used here is the binary cross entropy loss with sigmoid output.

$$Binary\ CE\ loss\ = \frac{1}{N}\sum\nolimits_{i=1}^{N} -\big(y_i * log(p_i) + (1 - y_i) * log(1 - p_i)\big)$$

*Equation 2. Binary Cross Entropy loss function*

## 2.9    Model Evaluation

We did model evaluation using model size and metrics such as Accuracy, Recall, and F1 scores. Inference times on five different test datasets have been calculated using different batch sizes using CPU as inference engine as images are processed sequentially on a CPU.

# 3 Analysis

## 3.1 Dataset description

There are 5307, 1326, and 715 images in the train, validation, and test dataset. The distribution of the classes is as follows. There is no significant class imbalance present in the dataset, as seen in Figure 3

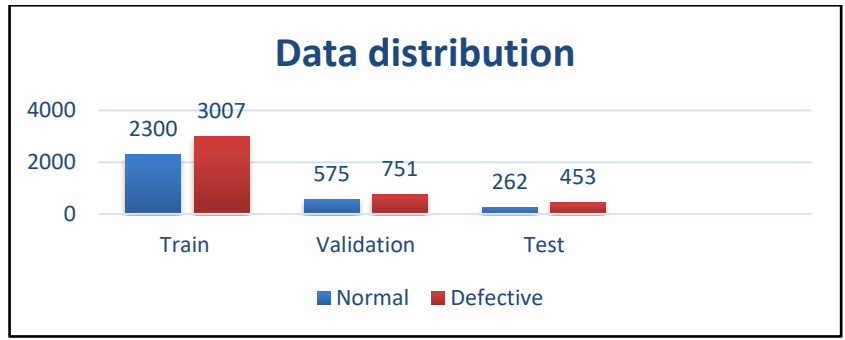

*Figure 2. Data distribution*

The visualization of the normal and the defective impellers in the RGB channel and grey scale images are shown in Figure 4 and Figure 5.

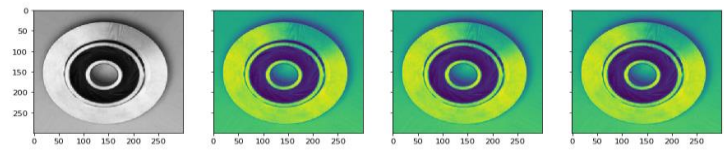

*Figure 3. RGB Images of the Impeller*

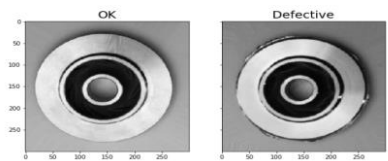

*Figure 4. Normal and Defective Impellers*

### 3.2    Compression of Models

The custom model has been built to reduce the number of training parameters. **Channel pruning** has been done in custom architectures. In contrast, the original network has been used in the case of transfer learning architectures as it is impossible to change these pre-trained networks' architecture. The models are saved using **keras's** *models.save* functionality which saves the model in **'.h5'** format, which is **more compressed** than the original **'.hdf5'** format. The network architectures are summarized in the table below.

| Models | Total params | Trainable params | Non trainable | Model Size MB |
|---|---|---|---|---|
| **Custom model – Normal** | **5,865** | **5,801** | **64** | **0.08** |
| MobileNetV2 | 2,260,546 | 2,226,434 | 34,112 | 9.52 |
| NasNet | 4,271,830 | 4,235,092 | 36,738 | 18.35 |
| Resnet50 | 23,591,810 | 23,538,690 | 53,120 | 94.89 |

*Table 1. Comparison of Model Sizes and Paramaters*

On analysis of the model architecture, it is found that custom models have the least file size (*0.08MB*) compared to the original architectures. The custom architectures have the least number of training parameters (5865) due to the shallow network.

### 3.3    Data Augmentation

As mentioned in the research methodology sections, different augmentation techniques have been tried out at model training and evaluated on the augmented test dataset. Some of the augmentation techniques tried are shown in Figure 6

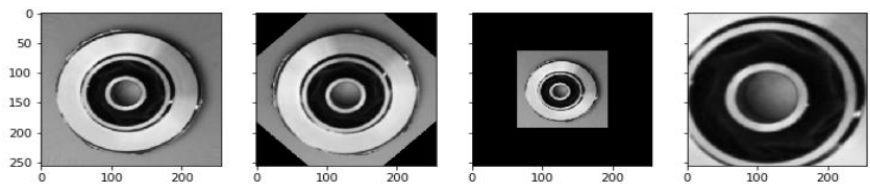

*Figure 5. Augmentations showing rotation and zooming*

## 3.4    Model architecture

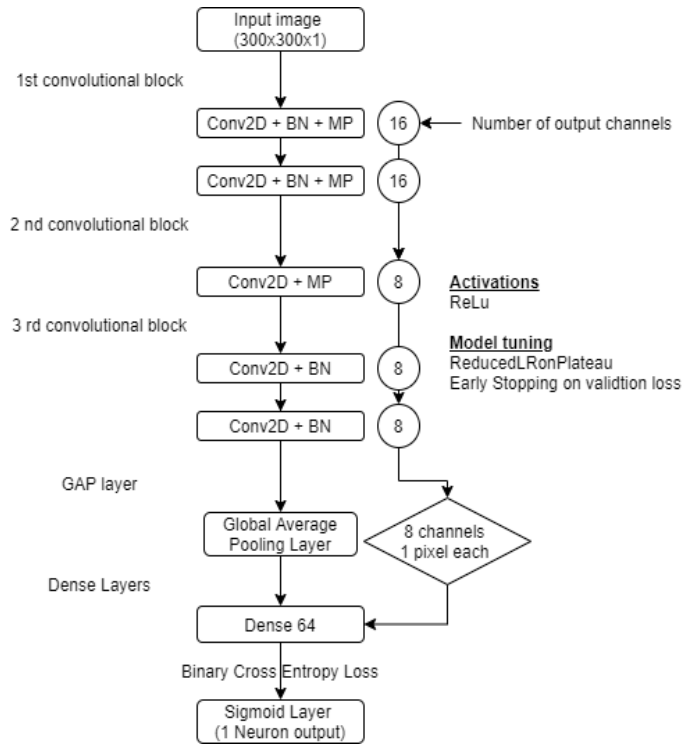

*Figure 6. Custom Architecture showing sequential pruning of number of channels*

**Custom Architecture.** The model architectures for the custom and the transfer learning models are explained in the succeeding paragraphs. Early stopping and ReducedLRonPlateau have been used at the time of model training to ensure no oscillation of the learning at the end of every epoch. The custom architecture consists of a sequential reduction in the number of channels from 16 to 8 and terminates at a global average pooling layer. The detailed architecture is shown in Figure 7.

**Transfer learning architectures.** The topmost layer of the transfer learning models is replaced with a custom softmax layer. The architectures of different pre-trained architectures like MobileNet, NasNet etc. are shown in Figures 8, 9, and 10.

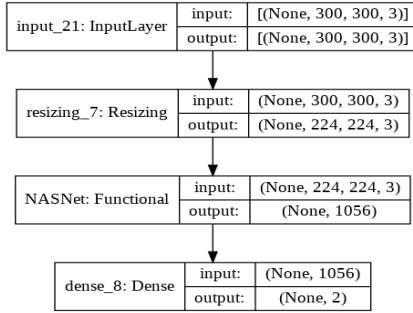

Figure 7. NasNet Architzecture

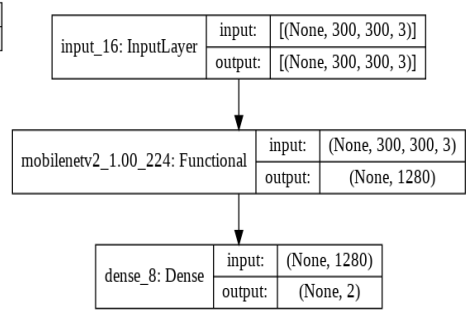

Figure 8. MobileNet Architecture

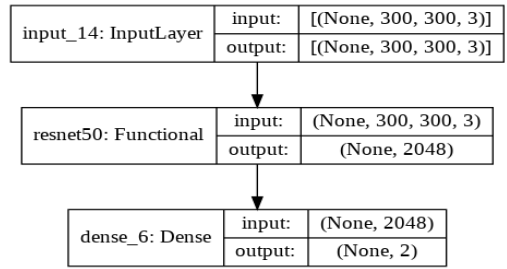

Figure 9. Resnet50 Architecture

# 4    Results

The results of the experiments have been explained in three sections; model evaluation, inference time analysis, and effect of augmentation on the best performing model (Custom Model).

## 4.1    Model Evaluation

Performance of various model architectures including custom model and transfer learning models are summarized in Table 2

| Models | Accu-racy | Re-call | F1 Score | Preci-sion | Total params | Size (MB) |
|---|---|---|---|---|---|---|
| Custom Model | **99.44** | **99.44** | **99.44** | **99.45** | **5,865** | **0.08** |
| MobileNetV2 | 98.04 | 98.04 | 98.05 | 98.14 | 2,260,546 | 9.52 |
| NasNet | 99.3 | 99.05 | 99.3 | 99.31 | 4,271,830 | 18.35 |
| Resnet50 | 99.16 | 99.16 | 99.16 | 99.16 | 23,591,810 | 94.89 |

*Table 2. Model evaluation Metrics*

Custom models achieve the highest evaluation metrics in terms of accuracy, recall, and F1 score. The number of parameters is also the least in the custom model, with a model size of just 0.08 MB. The inter-model performance ratios better represent the performance gains of the custom model on the total parameters and the model size in Table 3.

| Models | Parameter ratio | Model size ratio |
|---|---|---|
| Custom model | 1x | 1x |
| MobileNetV2 | 386x | 119x |
| NasNet | 728x | 229x |
| Resnet50 | 4022x | 1186x |

*Table 3. The ratio of parameters and model size*

## 4.2 Inference Time Analysis

The model trained on the standard test data set with batch size 32 has been evaluated on five different test datasets, as shown in table 2. From the below findings we can deduce that better the inference time lower the latency.

| Models | Test batch 1 | Test batch 10 | Test batch 50 | Test batch 100 | Test Batch 700 |
|---|---|---|---|---|---|
| Custom model | **0.0176** | **0.1344** | **0.3936** | **1.1853** | **12.6198** |
| MobileNetV2 | 0.0456 | 0.3151 | 1.2970 | 2.6959 | 21.7204 |
| NasNet | 0.0572 | 1.1835 | 3.5687 | 3.5167 | 26.6517 |
| Resnet50 | 0.1596 | 1.3304 | 3.5780 | 9.5807 | 76.7329 |

*Table 4. Inference Timing of various models*

**Lightweight faster custom architecture.** Custom models have the least inference times on a CPU compared to other transfer learning models when evaluated on all the different images (1,10,50,100,700), as shown in Table 2. Also, as seen in the trend curves, the order of inference times from the least to the maximum is

"**Custom model < MobileNetV2 < NasNet < Resnet50**"

The reduced inference time is attributed to the factors such as Channel pruning (Sequential reduction in the number of output channels), Single Class Output, and Decreased Kernel dimension due to grayscale input. The model size is also the least among all model architectures (0.08MB).

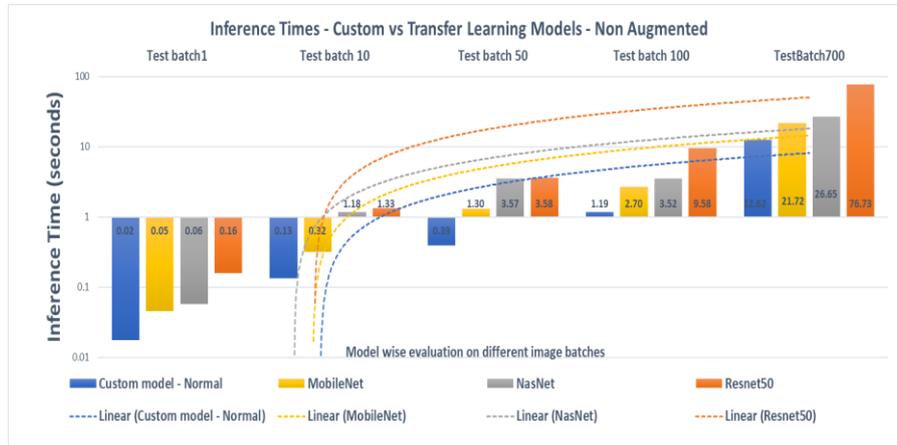

*Figure 10. CPU Inference Time for Models on a varying number of images*

The model's speed in terms of inference time is summarized in Table 5.

| Model vs inference times | Single Image | 10 images | 50 images | 100 images | 700 images |
|---|---|---|---|---|---|
| Custom model - Normal | 1x | 1x | 1x | 1x | 1x |
| MobileNetV2 | 2.58x | 2.34x | 3.30x | 2.27x | 1.72x |
| NasNet | 3.23x | 8.81x | 9.07x | 2.97x | 2.11x |
| Resnet50 | 9.02x | 9.90x | 9.09x | 8.08x | 6.08x |

*Table 5. Model inference Speeds*

Custom Models perform 6 to 9 times faster than the conventional Resnet architectures. Even among the lightweight architectures such as MobileNetV2 and NasNet, the Custom model performs anywhere between 2 to 8 times faster, indicating they are better suited for deploying this model on edge devices.

### 4.3 Augmentation on Custom model

The performance of the models is evaluated using F1 scores of models trained on both augmented and non-augmented datasets and by evaluating them on both the augmented/non augmented test datasets.

| Models | Acc | Re | F1 score | Pre |
|---|---|---|---|---|
| Custom model – Normal | 99.44 | 99.44 | 99.44 | 99.45 |
| Custom model – Augmented | 98.04 | 98.04 | 98.04 | 98.05 |
| Custom model (Aug) - Normal test | 99.16 | 99.16 | 99.16 | 99.17 |
| Custom model (Aug) - Augmented test | 98.18 | 98.18 | 98.17 | 98.2 |

*Table 6. Custom Model Evaluation Metrics*

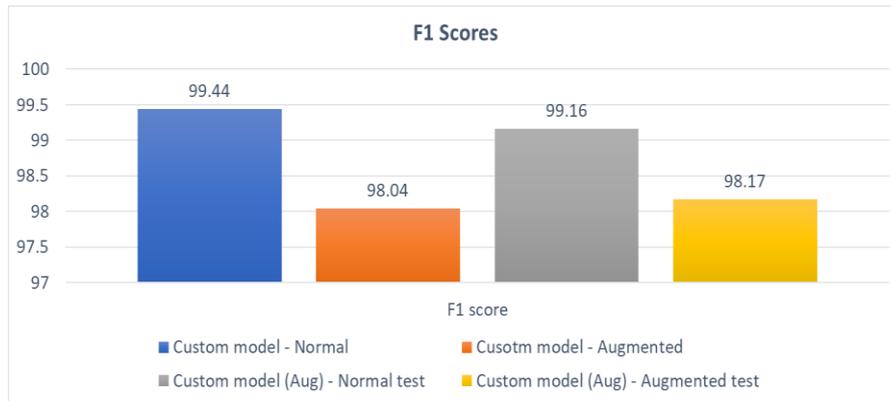

*Figure 11. Effect of augmentation on F1 scores of augmented/ non augmented models*

The F1 score of the custom model on the standard dataset is 99.44% and against the Augmented test dataset is 98.04%. Though the models trained on the augmented dataset show a slightly lower F1 score on the standard dataset (99.16% vs. 99.44%), they show higher performance on the augmented test dataset (98.17% vs. 98.04%). Hence, augmentation techniques, though they do not significantly improve accuracy, may still be chosen over non-augmented models as they are more robust and not susceptible to changes in the training dataset.

## 5    Conclusion

The architectures of transfer learning models that use pre-trained weights are typically larger. Even if they have a high level of accuracy across a large number of datasets, they may not be ideal for use on computationally less powerful devices. Compared to transfer learning models, custom models with a model size of just 0.08MB could perform better in terms of F1 scores on the given dataset. Inference times on various images showed that the models performed much faster than transfer learning models. The performance of augmented models was comparable to that of non-augmented models; however, these models may be preferred because they are more robust and may perform well on other test datasets. In terms of implementation or deployment on smaller devices, the study establishes the superiority of custom architectures over large scale or pre-trained models.